\documentclass{article} 
\usepackage{iclr2027_conference,times}

\usepackage{amsmath,amsfonts,bm}

\def\eqref#1{equation~\ref{#1}}

\def\1{\bm{1}}

\DeclareMathAlphabet{\mathsfit}{\encodingdefault}{\sfdefault}{m}{sl}
\SetMathAlphabet{\mathsfit}{bold}{\encodingdefault}{\sfdefault}{bx}{n}

\usepackage{hyperref}
\usepackage{url}

\title{Can AI Make Money in Crypto? Measuring the Gap from Backtests to Real Markets}

\author{
\textbf{
Xingtong Yu\textsuperscript{1},
Jiarun Zhou\textsuperscript{2},
Guanlin Ding\textsuperscript{2},
Wenkang Wei\textsuperscript{2},
Jiarui Liu\textsuperscript{3},
Chang Zhou\textsuperscript{2},
}\\
\textbf{
~Fangzhou Ge\textsuperscript{1},
Chenyi Xu\textsuperscript{1},
Xikun Zhang\textsuperscript{4},
Renqiang Luo\textsuperscript{3},
Jie Zhang\textsuperscript{5},
Hong Cheng\textsuperscript{1},}\\
\textbf{
~Xinming Zhang\textsuperscript{2},
Hui Zhang\textsuperscript{2}$^{*}$,
Yuan Fang\textsuperscript{6}$^{*}$
}\\
~The Chinese University of Hong Kong\textsuperscript{1},
University of Science and Technology of China\textsuperscript{2},\\
~Jilin University\textsuperscript{3},
RMIT University\textsuperscript{4},
Infplane Computing Lab\textsuperscript{5},
Singapore Management University\textsuperscript{6}\\
~\texttt{xtyu@se.cuhk.edu.hk},
\texttt{fzhh@ustc.edu.cn},
\texttt{yfang@smu.edu.sg}
}

\newcommand{\stitle}[1]{\vspace{0mm} \noindent {\bf #1}} 
\newcommand{\method}[1]{\textsc{#1}}
\usepackage[most]{tcolorbox}
\usepackage{xcolor}
\newtcolorbox{promptbox}[1]{
  colback=gray!5,
  colframe=gray!60,
  boxrule=0.5pt,
  arc=2pt,
  left=6pt,
  right=6pt,
  top=5pt,
  bottom=5pt,
  fonttitle=\bfseries,
  title=#1,
  breakable
}

\iclrfinalcopy 

\usepackage{booktabs}
\usepackage{multirow}
\usepackage{graphicx}
\usepackage{makecell}
\usepackage{wrapfig}

\begin{document}

\maketitle

\begin{abstract}
AI-based trading methods have rapidly evolved from machine learning and reinforcement learning to large language models (LLMs) and trading agents, yet their performance is still predominantly assessed through historical backtesting. Such evaluations provide limited evidence of whether a method can generalize to unseen future markets or whether its backtested performance can be sustained in  realistic trading frictions (e.g., latency, slippage, liquidity constraints, and market impact). 
We present a unified benchmark that evaluates representative machine learning, reinforcement learning, LLM-based, and agent-based trading methods in cryptocurrency markets through three progressively more realistic stages: historical backtesting, prospective exchange-based paper trading, and real-money live trading. These stages jointly increase temporal realism by moving from historical to unseen future markets, and execution realism by moving from offline simulation toward live trading. This protocol enables us to quantify the backtest-to-realization gap, identify when performance begins to deteriorate, and compare how this gap differs across major classes of AI trading methods. We further provide a unified open-source system supporting all three evaluation stages, together with a public platform that continuously updates benchmark results. Code is available at \url{https://github.com/Starlien95/Awesome-TradingAI}.
\end{abstract}

\section{Introduction}
Artificial intelligence has been increasingly applied to financial trading, giving rise to a wide range of methods based on machine learning (ML) \citep{fischer2018deep,zhang2019deeplob}, reinforcement learning (RL) \citep{zong2024macrohft}, large language models (LLMs) \citep{yang2023fingpt}, and trading agents \citep{xiao2024tradingagents}. These methods differ in how they process market information and make trading decisions, yet they are commonly evaluated in historical backtests or simulated environments \citep{zong2024macrohft,zhang2024finagent,yu2024finmem,xiao2024tradingagents}. Such evaluations can show how a method performs on past market data, but they provide limited evidence of whether that performance will persist on unseen future data or under realistic trading conditions. 

Answering this question requires considering two dimensions of evaluation realism. The first is \emph{temporal realism}: a method that performs well on historical data may not generalize to future markets as market conditions change over time. This concern becomes more pronounced when models are extensively tuned and selected using historical data. The second is \emph{execution realism}: offline evaluation simplifies how trading decisions are translated into actual orders and realized returns, whereas deployed systems must interact with an exchange under practical execution constraints such as latency, slippage, liquidity, and market impact. Together, these two dimensions help distinguish whether performance degradation arises from poor generalization to future markets or from the execution of trading decisions in live markets.

Current benchmarks do not yet provide this complete view. Many still rely on historical or pre-collected datasets \citep{liu2022finrlmeta,sun2023trademaster,xie2024finben,li2025investorbench,chen2025stockbench}. More recent benchmarks improve temporal realism by evaluating systems on newly arriving market data \citep{li2025deepfund,qian2026whenagents,yu2025livetradebench,fan2025aitrader}. However, they still do not evaluate methods through a common staged protocol spanning historical backtesting, exchange-based paper trading, and real-money live trading. Moreover, existing benchmarks are typically restricted to a specific methodological family, focusing on only RL \citep{liu2022finrlmeta,sun2023trademaster} or LLM-based trading systems \citep{xie2024finben,li2025investorbench,li2025deepfund,qian2026whenagents,yu2025livetradebench,fan2025aitrader}. This makes it difficult to compare different classes of AI trading methods under the same evaluation protocol or to determine how the \emph{backtest-to-realization gap} differs across them.

To address these gaps, we introduce a unified benchmark and open-source system for evaluating AI trading methods in cryptocurrency markets. Our benchmark covers representative methods across ML, RL, LLM-based, and agentic trading algorithms, and evaluates them consistently using a three-stage protocol comprising historical backtesting, prospective exchange-based paper trading, and live trading. By moving from historical to unseen future markets and from offline simulation to exchange-based or live execution, these stages progressively increase both temporal and execution realism. This staged evaluation enables us to measure the backtest-to-realization gap and compare how trading performance changes across different classes of AI trading methods. We integrate all methods into a common execution and evaluation framework and release the full codebase together with a public dashboard that continuously updates results across the three stages.

In summary, we make the following contributions.
(1) We introduce a unified benchmark that evaluates representative ML, RL, LLM-based, and agentic trading methods in cryptocurrency markets under a common evaluation protocol.
(2) We design a staged evaluation protocol spanning historical backtesting, exchange-based paper trading, and real-money live trading, progressively increasing both temporal and execution realism.
(3) We systematically characterize the backtest-to-realization gap across different classes of AI trading methods.
(4) We release a complete open-source system that supports all three evaluation stages, together with implementations of the evaluated methods and a public platform that continuously updates benchmark results.

\section{Related Work}

\stitle{AI for trading.}
AI-based trading methods span several major paradigms. ML methods typically learn predictive signals from historical market data and use these signals to construct trading strategies \citep{fischer2018deep,zhang2019deeplob}. Recent work further improves prediction under temporal distribution shifts, noisy financial data, and cross-asset dependencies \citep{lin2021tra,du2021adarnn,wu2021tcts,zhang2020doubleensemble,xu2021igmtf}. RL instead learns trading policies directly from interaction rewards, with recent methods introducing hierarchical decision making and market-aware adaptation for high-frequency trading \citep{qin2024earnhft,zong2024macrohft}. More recently, LLM-based methods incorporate textual information and reasoning into trading decisions \citep{yang2023fingpt}, while trading agents further introduce mechanisms such as memory, reflection, and multi-agent collaboration \citep{yu2024finmem,zhang2024finagent,xiao2024tradingagents}. 

\stitle{Benchmarks for AI trading.}
Existing AI trading benchmarks mainly differ in evaluation realism and methodological scope. Earlier benchmarks such as FinRL-Meta \citep{liu2022finrlmeta} and TradeMaster \citep{sun2023trademaster} standardize trading environments and evaluation protocols, but primarily evaluate reinforcement-learning methods on historical market data. FinBen \citep{xie2024finben}, InvestorBench \citep{li2025investorbench}, and StockBench \citep{chen2025stockbench} extend evaluation toward LLM-based financial decision-making, but remain centered on language models or agents rather than comparing different classes of trading methods. More recent benchmarks, including DeepFund \citep{li2025deepfund}, Agent Market Arena \citep{qian2026whenagents}, LiveTradeBench \citep{yu2025livetradebench}, and AI-Trader \citep{fan2025aitrader}, further improve temporal realism by evaluating systems on newly arriving market data. However, these benchmarks remain centered on LLM-based systems and do not evaluate heterogeneous trading methods under the same staged protocol spanning historical backtesting, exchange-based paper trading, and real-money trading. Our benchmark complements these efforts by covering ML, RL, LLM-based methods, and trading agents within this common protocol.
\section{Benchmarking Protocols}\label{sec.setting}

\subsection{Evaluated Methods}
We evaluate 32 AI trading methods spanning ML, RL, LLM-based trading, and trading agents, together with Bitcoin buy and hold as a passive market baseline. The evaluated methods cover predictive models, sequential decision-making policies, direct LLM-based trading, and agent-based systems with memory and multi-agent collaboration. Detailed descriptions are provided in Appendix~\ref{app.methods}.
For reproducibility, we build each method from its publicly released codebase when available and adapt it to our unified trading framework. Model selection is conducted exclusively through historical backtesting over model hyperparameters and trading configurations, including trading windows, position limits, and leverage. For each method, we select the configuration with the best validation performance and keep it fixed in later stages. Detailed configurations are provided in Appendix~\ref{app.hyperparams}. To control randomness, we use a fixed random seed for non-LLM methods, while each historical backtest for LLM-based methods and trading agents is run three times and averaged.

\subsection{Trading Task Formulations}
We consider a tradable universe of ten large-cap cryptocurrencies, which serves as the common set of candidate assets throughout the benchmark. We provide the details of the assets in Appendix~\ref{app.assets}. Depending on the trading formulation, a method may operate on either a predefined asset or autonomously selected assets within this universe. At each decision step $t$, a trading method $m$ uses the market information available up to that time to determine \emph{what to trade} and \emph{how to trade it}. We summarize this process as
\begin{equation} 
\mathcal{S}_{t,m} = S_m(\mathcal{I}_{t,m}), \qquad a_{t,m} = \Pi_m(\mathcal{I}_{t,m}, \mathcal{S}_{t,m}), 
\end{equation}
where $\mathcal{I}_{t,m}$ denotes the market information available to method $m$, such as historical open, high, low, close, and volume (OHLCV) data\citep{park2022stock}, derived market factors \citep{yang2020qlib}, portfolio states, or textual information related to the traded assets. $\mathcal{S}_{t,m}$ denotes the assets selected for trading, such as selected cryptocurrencies in a multi-asset strategy or a predefined asset. The action $a_{t,m}$ specifies how the selected assets should be traded, such as buying, selling, or holding, choosing a long or short direction, determining the position size, or specifying risk-control instructions such as stop-loss and target prices.

\stitle{Machine learning.} 
For ML methods, $\mathcal{I}_{t,m}$ consists of 52 market factors (Alpha52) extracted from historical OHLCV data. We construct these factors from the raw market data and use them as the sole inputs to the ML models. Each model then predicts the future return of every candidate asset over a predefined horizon. Following previous work \citep{yang2020qlib}, we then rank the candidate assets according to their predicted returns, select the top-$k$ as $\mathcal{S}_{t,m}$, and construct a long-only portfolio over the selected assets. 

\stitle{Reinforcement learning.} 
For RL methods, the trading asset is predefined rather than selected dynamically from the candidate asset universe. The RL agents use the same Alpha52 market factors as the ML methods as input. Instead of predicting future returns and ranking assets, the RL policy directly determines how the position in the predefined asset should be adjusted based on the current market state. We use a long-only setting and preserve the original action formulation of each RL method when mapping its decisions to executable orders.

\stitle{LLM-based methods and trading agents.} 
For LLM-based methods and trading agents, we construct method-specific prompts from the market data and asset-related textual information available for the candidate assets. Depending on the method, the prompt either specifies a target asset or provides information about a set of candidate assets from which the method selects what to trade. The method then determines the corresponding trading actions. Detailed prompt templates and input configurations are provided in Appendix~\ref{app.hyperparams}. Specifically, \method{DeepSeek} and \method{Qwen} output the selected asset together with a buy, sell, or hold decision, trading amount, stop-loss price, and target price. \method{FinGPT} follows a long-only strategy and determines which asset to trade and whether to buy or sell the available position. \method{FinMem} selects an asset and chooses among three discrete actions: taking a full long position, taking a full short position, or maintaining the current position. \method{FinAgent} and \method{TradingAgents} output the selected asset, a long or short direction, and the corresponding position size, with the position capped at 15\% of the account value following their corresponding trading protocols.

\subsection{Evaluation Protocol}
We evaluate trading methods through three stages: historical backtesting, exchange-based paper trading, and real-money trading. For any method advanced to a later stage, its trading formulation and selected configuration are kept fixed, while the market timeline and execution environment become progressively closer to actual deployment.

\stitle{Evaluation metrics.}
We evaluate trading outcomes using
 well-established performance and risk metrics in quantitative finance. Specifically, \emph{Cumulative return} measures the net change in portfolio value over the evaluation period; \emph{Jensen's alpha} \citep{jensen1968performance} measures the excess return of a strategy relative to that implied by its exposure to market risk, with Bitcoin Buy\&Hold adopted as the market benchmark in our evaluation; \emph{Sharpe ratio} \citep{sharpe1966mutual} measures return relative to the total risk taken by the strategy, providing a risk-adjusted measure of performance; \emph{Maximum drawdown} \citep{magdon2004maximum} captures the largest peak-to-trough decline in portfolio value and reflects the severity of sustained losses; \emph{Volatility} \citep{fabozzi2008portfolio} measures the variability of portfolio returns and characterizes the overall level of return risk.

\stitle{Execution and transaction costs.}
We use the OKX cryptocurrency exchange as the execution environment for both paper trading and live trading. All trades are submitted as taker orders to facilitate faster execution against available market liquidity. Long positions are implemented through spot trading and incur a transaction fee of 0.10\% per trade, while short positions are implemented through perpetual futures and incur a taker fee of 0.05\% per trade. These fee rates follow the OKX taker fee schedule used in our experiments. We apply the same transaction costs in historical backtesting to maintain consistent trading-cost assumptions across evaluation stages.

\stitle{Historical backtesting.} 
We first evaluate each method on historical market data collected from the OKX cryptocurrency exchange. We use data from 2021-01-01 to 2023-12-31 for training, from 2024-01-01 to 2024-12-31 for validation, and from 2025-01-01 to 2025-12-31 for testing. The validation period is used for model and trading-configuration selection, while all reported backtesting results are computed on the held-out 2025 test period. At each decision step, the method uses only information that would have been available at that time, and its trades are simulated using the corresponding historical market prices. This stage measures how each method would have performed in past markets under simulated trading, providing a historical baseline before prospective evaluation.

\stitle{Paper trading.}
We then advance the best-performing methods from historical backtesting to prospective evaluation on newly arriving market data in an exchange-hosted paper-trading environment. Paper trading is conducted from 2026-06-04 to 2026-09-21. To retain representation across methodological families, we select the top-performing methods from each category based on their backtesting results. Unlike backtesting, future market movements are unknown when decisions are made, and orders are submitted to the exchange environment using simulated capital. This stage therefore tests whether strong backtesting performance can persist in unseen future markets under more realistic execution conditions.

\stitle{Live trading.}
Finally, we advance the best-performing methods from exchange-based paper trading to live trading on the exchange. Live trading is conducted from 2026-08-24 to 2026-09-21. Specifically, we select the five methods with the highest Jensen's alpha during paper trading and deploy them with real capital. The methods continue to make decisions prospectively, but their orders are now executed in the live market and generate realized trading outcomes. This stage tests whether strong paper-trading performance can persist when the methods are deployed with real capital.

\section{Empirical Results and Analysis}
Our experiments follow the three-stage evaluation protocol introduced in Sect.~\ref{sec.setting}. We first evaluate all methods through historical backtesting, then select representative high-performing methods from each methodological category for exchange-based paper trading, and finally deploy the top four methods with the highest paper-trading alpha as of 2026-08-23 to real-money trading.

\subsection{Historical Backtesting}
\textbf{RQ1:} \textit{How do different classes of AI trading methods perform under historical backtesting?}

\begin{table*}[t]
\centering
\scriptsize
\caption{Backtest performance of the evaluated trading methods from 2025-01-01 to 2025-12-31.}
\label{tab:best_cr_results}
\resizebox{\linewidth}{!}{
\begin{tabular}{lccccc}
\toprule
\textbf{Method} &
\textbf{Cum. Ret. (\%)$\uparrow$} &
\textbf{Alpha (\%)$\uparrow$} &
\textbf{Sharpe$\uparrow$} &
\textbf{Max DD (\%)$\downarrow$} &
\textbf{Vol. (\%)$\downarrow$} \\
\midrule

\method{Buy\&Hold} & -7.63 & 0.00 & -0.18 & 33.01 & 41.74 \\

\midrule
\method{LightGBM} & 8.57 & 13.70 & 0.31 & 52.05 & 42.14 \\
\method{CatBoost} & 21.46 & 25.61 & 0.57 & 61.93 & 58.07 \\
\method{Linear} & 21.08 & 25.87 & 0.63 & 44.59 & 51.30 \\
\method{XGBoost} & 41.93 & 53.60 & 1.08 & 42.42 & 59.50 \\
\method{MLP} & 30.33 & 39.15 & 1.08 & 35.24 & 43.06 \\
\method{TabNet} & 34.52 & 45.34 & 1.27 & 32.67 & 41.59 \\
\method{TCN} & \underline{62.94} & 86.10 & \textbf{1.62} & \textbf{29.10} & 59.48 \\
\method{GRU} & 29.66 & 37.40 & 1.02 & 38.14 & 44.49 \\
\method{LSTM} & 32.90 & 39.33 & 0.82 & 44.07 & 61.76 \\
\method{SFM} & 18.31 & 21.60 & 0.68 & 43.95 & 41.37 \\
\method{GeneralPtNN} & 16.44 & 20.54 & 0.55 & 34.04 & 45.80 \\
\method{ALSTM} & 21.17 & 26.30 & 0.70 & 51.26 & 46.48 \\
\method{TFT} & 10.53 & 13.93 & 0.38 & 50.30 & 42.47 \\
\method{Transformer} & 5.32 & 8.90 & 0.20 & 41.11 & 41.64 \\
\method{KRNN} & 9.78 & 15.32 & 0.33 & 35.75 & 45.88 \\
\method{Localformer} & 24.38 & 30.63 & 0.68 & 51.19 & 55.30 \\
\method{TRA} & 25.08 & 31.39 & 0.54 & 52.94 & 70.66 \\
\method{AdaRNN} & 19.41 & 26.87 & 0.46 & 45.59 & 64.38 \\
\method{TCTS} & 10.92 & 16.53 & 0.40 & 50.06 & 41.56 \\
\method{ADD} & -1.37 & 1.71 & -0.05 & 52.05 & 41.72 \\
\method{DoubleEnsemble} & 37.48 & 47.97 & 1.04 & 48.39 & 55.52 \\
\method{GATs} & \textbf{71.70} & \underline{98.03} & \underline{1.57} & 33.97 & 70.08 \\
\method{Sandwich} & 23.16 & 28.66 & 0.62 & 35.89 & 57.12 \\
\method{IGMTF} & 1.72 & 5.54 & 0.06 & 58.33 & 44.80 \\

\midrule
\method{MacroHFT} & -10.07 & -3.10 & 0.04 & 45.18 & 50.58 \\
\method{EarnHFT} & -38.57 & -76.96 & -1.75 & 45.75 & 44.43 \\

\midrule
\method{DeepSeek} & -43.30 & \textbf{278.44} & -0.22 & 96.85 & 261.12 \\
\method{Qwen} & -68.62 & 38.38 & -0.62 & 82.85 & 188.58 \\
\method{FinGPT} & -15.15 & -7.03 & 0.03 & 45.13 & 60.08 \\

\midrule
\method{FinMem} & -20.76 & -19.30 & -0.78 & 41.08 & \textbf{29.95} \\
\method{FinAgent} & -41.06 & -48.14 & -1.61 & 45.51 & \underline{32.78} \\
\method{TradingAgents} & -9.16 & -3.36 & -0.25 & \underline{31.79} & 38.02 \\

\bottomrule
\end{tabular}}
\end{table*}

\stitle{Backtest performance.} 
We report the historical backtesting performance of all evaluated methods during 2025 in Table~\ref{tab:best_cr_results}. 
We first observe a clear performance difference across methodological families. Among the 24 ML methods, 23 achieve positive cumulative returns, whereas none of the evaluated RL-, LLM-, or agent-based methods is profitable over the same period. Among the latter groups, \method{TradingAgents} performs closest to the \method{Buy\&Hold}, while several other methods incur substantially larger losses. 
Second, strong family-level performance does not imply consistently strong individual methods. Within ML, cumulative returns range from $-1.37\%$ to $71.70\%$, despite the methods using the same market factors and trading protocol. \method{GATs} and \method{TCN} achieve the strongest results, while several other ML methods deliver only modest gains. Thus, substantial performance variation remains even under a common trading formulation. 
Third, different methods exhibit distinct strengths across evaluation dimensions. \method{GATs} achieves the highest cumulative return, whereas \method{TCN} obtains the highest Sharpe ratio and the lowest maximum drawdown. \method{DeepSeek} attains the highest alpha despite a negative final return, reflecting a highly volatile leveraged trading trajectory with substantial intermediate gains and losses. Its high volatility and maximum drawdown further illustrate that a strong value on one metric does not necessarily correspond to robust overall trading performance. These results show that no single metric can fully characterize a method's trading performance, motivating evaluation across both profitability and risk.

\subsection{Exchange-Based Paper Trading}

\textbf{RQ2:} \textit{How much of the backtested performance persists in prospective paper trading?}

\begin{figure}[t]
  \centering
  \begin{minipage}[t]{0.49\linewidth}
    \centering
    \includegraphics[width=0.88\linewidth]{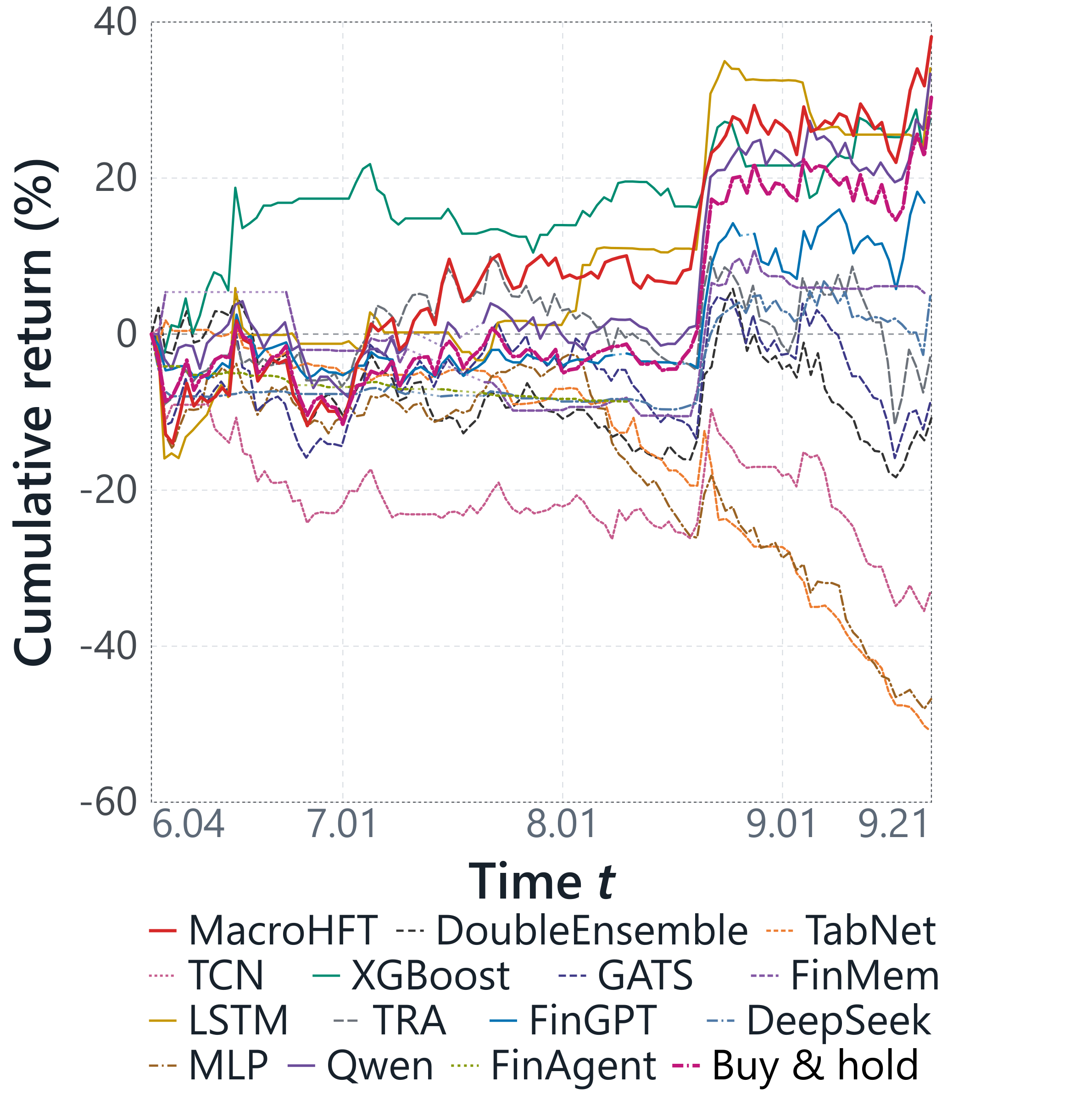}
    \vspace{-5mm}
    \caption{Cumulative returns of paper trading.}
    \label{fig:paper-trading}
  \end{minipage}\hfill
  \begin{minipage}[t]{0.49\linewidth}
    \centering
    \includegraphics[width=0.88\linewidth]{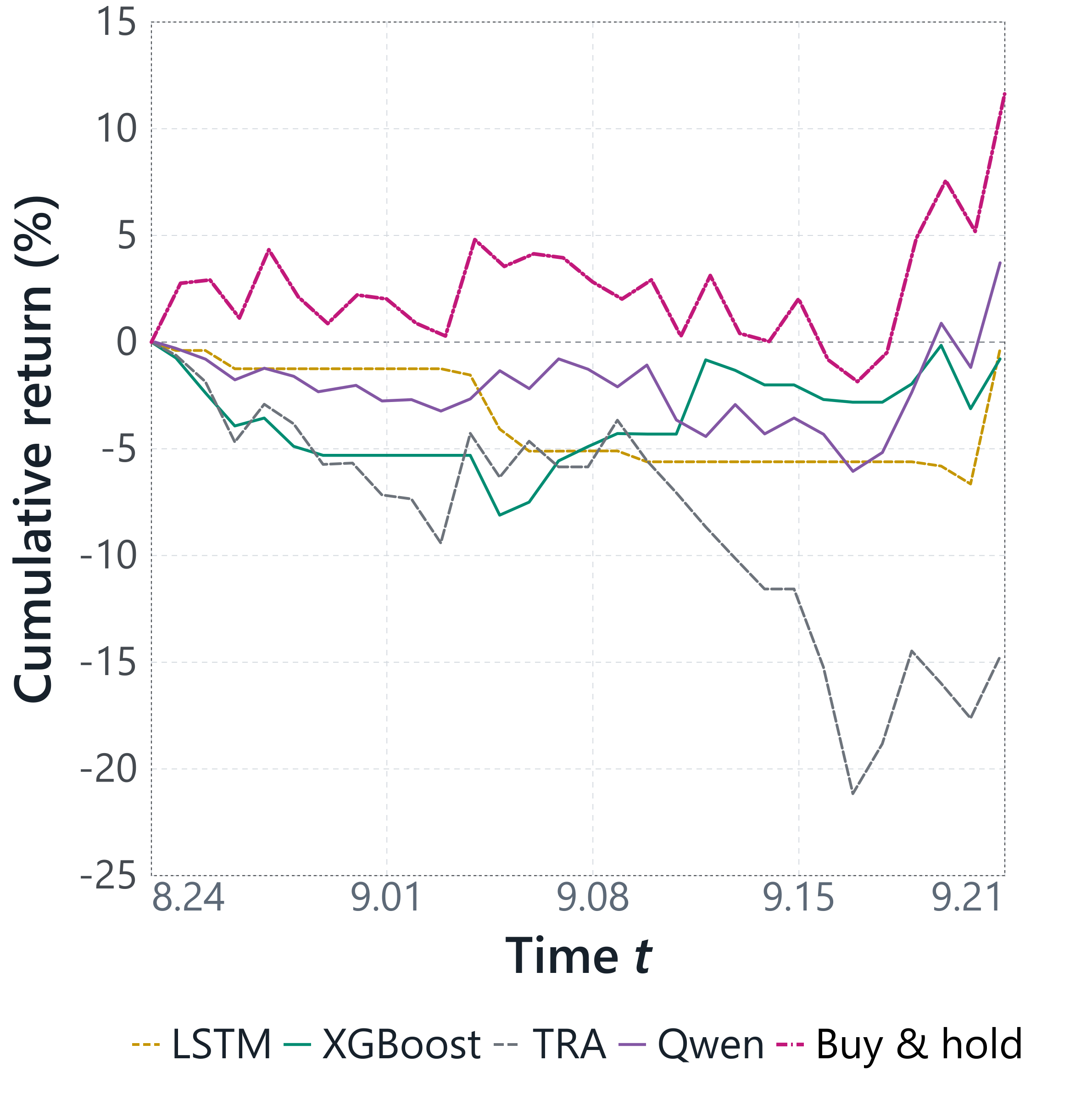}
    \vspace{-5mm}
    \caption{Cumulative returns of live trading.}
    \label{fig:live-trading}
  \end{minipage}
\end{figure}

\begin{table*}[t]
\centering
\scriptsize
\setlength{\tabcolsep}{3pt}
\caption{Backtest (BT) and paper trading (PT) performance.}
\label{tab:backtest-papertrading-results}
\resizebox{\linewidth}{!}{
\begin{tabular}{lcccccccccc}
\toprule
\textbf{Method} &
\multicolumn{2}{c}{\textbf{Cum. Ret. (\%)$\uparrow$}} &
\multicolumn{2}{c}{\textbf{Alpha (\%)$\uparrow$}} &
\multicolumn{2}{c}{\textbf{Sharpe$\uparrow$}} &
\multicolumn{2}{c}{\textbf{Max DD (\%)$\downarrow$}} &
\multicolumn{2}{c}{\textbf{Vol. (\%)$\downarrow$}} \\
\cmidrule(lr){2-3}
\cmidrule(lr){4-5}
\cmidrule(lr){6-7}
\cmidrule(lr){8-9}
\cmidrule(lr){10-11}
& \textbf{BT} & \textbf{PT}
& \textbf{BT} & \textbf{PT}
& \textbf{BT} & \textbf{PT}
& \textbf{BT} & \textbf{PT}
& \textbf{BT} & \textbf{PT} \\
\midrule

\method{Buy\&Hold}
& \multicolumn{2}{c}{30.35}
& \multicolumn{2}{c}{0.00}
& \multicolumn{2}{c}{1.88}
& \multicolumn{2}{c}{13.56}
& \multicolumn{2}{c}{41.78} \\
\midrule

\method{XGBoost}
& \underline{33.95} & 27.53 & \textbf{64.23} & \underline{45.32}
& \textbf{2.90} & \underline{2.59} & 12.01 & 9.56 & 37.88 & 32.15 \\

\method{MLP}
& -45.56 & -46.79 & -256.24 & -251.80 & -4.69 & -4.77
& 49.63 & 48.97 & 45.52 & 44.03 \\

\method{TabNet}
& -54.37 & -50.80 & -257.37 & -237.74 & -10.30 & -8.34
& 56.27 & 52.72 & 24.91 & 27.67 \\

\method{TCN}
& -35.24 & -33.20 & -191.51 & -188.46 & -3.42 & -3.24
& 37.76 & 37.45 & 42.30 & 42.81 \\

\method{LSTM}
& 24.09 & 34.08 & 18.49 & 35.62 & 1.40 & 1.68
& 17.00 & 17.00 & 55.03 & 56.53 \\

\method{TRA}
& -3.84 & -2.98 & -80.01 & -82.15 & -0.04 & -0.01
& 19.05 & 20.53 & 44.59 & 46.01 \\

\method{DoubleEnsemble}
& -18.96 & -10.78 & -118.59 & -97.65 & -2.04 & -0.98
& 29.60 & 23.91 & 35.19 & 40.13 \\

\method{GATs}
& -15.88 & -8.79 & -147.79 & -117.05 & -1.24 & -0.56
& 28.28 & 22.80 & 49.15 & 50.42 \\

\midrule

\method{MacroHFT}
& \textbf{41.01} & \textbf{38.12} & \underline{30.09} & 24.20
& \underline{2.45} & 2.32 & 16.69 & 16.68 & 47.80 & 47.40 \\

\midrule

\method{DeepSeek}
& 7.35 & 9.53 & 1.61 & -0.91 & 1.43 & 1.32
& \textbf{3.31} & \underline{8.83}
& \textbf{12.40} & \underline{25.38} \\

\method{Qwen}
& 11.35 & \underline{37.77} & 15.66 & \textbf{53.15}
& 1.18 & \textbf{2.94} & \underline{8.49} & 11.83
& \underline{20.77} & 38.99 \\

\method{FinGPT}
& 20.04 & 16.85 & 14.15 & 3.51 & 1.90 & 1.62
& 9.41 & 8.97 & 35.38 & 36.13 \\

\midrule

\method{FinMem}
& -6.63 & 5.28 & -20.98 & -2.93 & -0.51 & 0.74
& 20.43 & 15.10 & 34.23 & 28.68 \\

\method{FinAgent}
& 0.20 & 1.73 & -10.08 & -15.28 & -0.21 & 0.43
& 15.31 & \textbf{7.88} & 22.50 & \textbf{16.56} \\


\bottomrule
\end{tabular}}
\vspace{2pt}
\end{table*}

\begin{figure}[t]
    \centering
    \includegraphics[width=\linewidth]{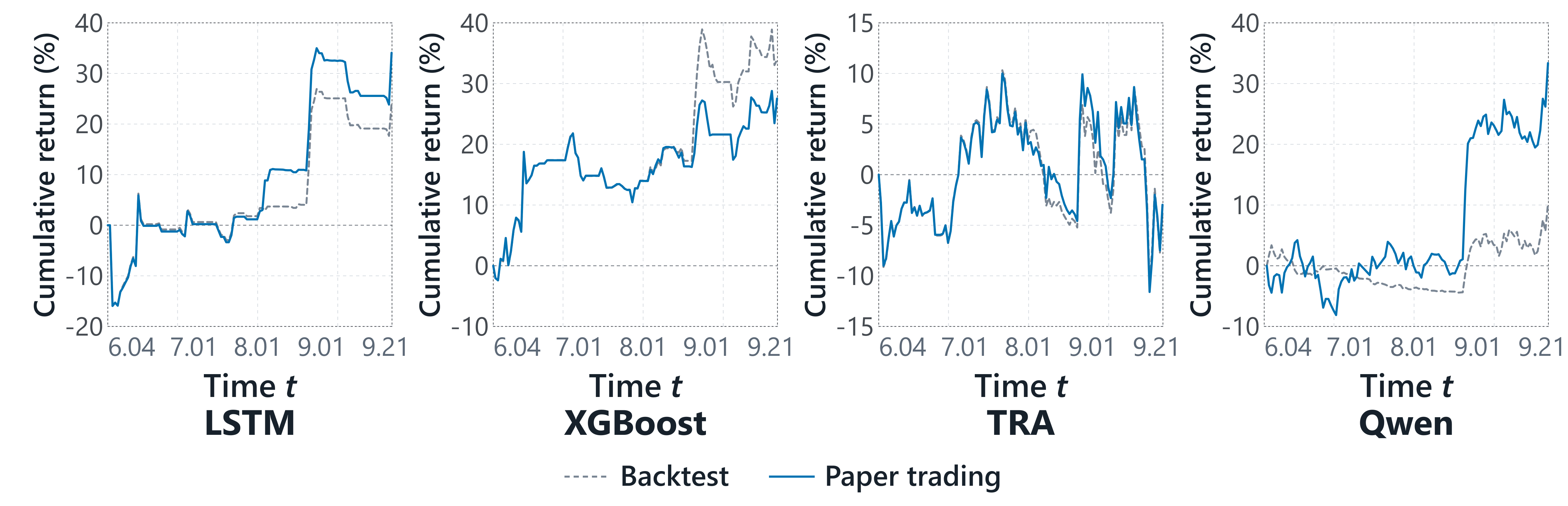}
    \vspace{-8mm}
    \caption{Cumulative returns of backtesting and paper trading.}
    \label{fig:pt-bt}
\end{figure}

\stitle{Paper-trading performance.} 
We report the cumulative-return curves of the selected methods during paper trading in Fig.~\ref{fig:paper-trading}, and summarize their final performance in Table~\ref{tab:backtest-papertrading-results}. These results reveal a markedly different pattern from historical backtesting.
We first observe a striking reversal across methodological families. Among the methods advanced to paper trading, all eight ML methods were profitable in the 2025 historical backtest, whereas none of the six selected RL-, LLM-, or agent-based methods were. In paper trading, this pattern nearly reverses: only two of the eight ML methods remain profitable, while all six methods from the other families achieve positive returns.
Second, this reversal is also reflected in the typical performance within each group. The median return of the selected ML methods drops from $36.0\%$ in historical backtesting to $-9.8\%$ in paper trading, whereas the median return of the selected non-ML methods rises from $-29.9\%$ to $13.2\%$. These results indicate that the relative advantage of different trading paradigms can change substantially when evaluation moves from historical to unseen future markets.

\stitle{Backtest-to-paper-trading comparison.}
We further compare backtesting and paper trading over the same market period from 2026-06-04 to 2026-09-21. Table~\ref{tab:backtest-papertrading-results} reports the results under both settings, while Fig.~\ref{fig:pt-bt} compares their cumulative-return curves for representative methods.
We first observe substantial discrepancies between backtesting and paper trading even for the same method over the same market period, with the cumulative-return gap reaching $26.42$ percentage points (pp) for \method{Qwen}. Notably, paper trading does not systematically underperform backtesting: 10 of the 14 evaluated methods achieve higher cumulative returns in paper trading.
Second, the discrepancy extends beyond cumulative return to the broader risk-return profile. Across the 14 methods, the average absolute backtest-to-paper-trading gap reaches $6.25$ pp in cumulative return and $14.05$ pp in alpha, together with differences of $0.62$ in Sharpe ratio, $2.98$ pp in maximum drawdown, and $4.53$ pp in volatility. Thus, backtesting can misestimate not only profitability but also the risk characteristics of a trading method.
Third, these discrepancies are large enough to change practical conclusions about individual methods. \method{FinMem}, for example, changes from a negative return in backtesting to a positive return of $5.28\%$ in paper trading. \method{Qwen} moves from substantially underperforming the \method{Buy\&Hold} baseline to outperforming it, whereas \method{XGBoost} shows the opposite pattern. These reversals show that backtesting alone can lead to qualitatively different conclusions about whether a method is profitable or competitive with the market.

\begin{wraptable}{r}{0.5\columnwidth}
\vspace{-8pt}
\centering
\scriptsize
\caption{Paper trading execution discrepancies.}
\label{tab:papertrading_execution_gap}
\setlength{\tabcolsep}{2.5pt}
\resizebox{0.5\columnwidth}{!}{
\begin{tabular}{lccc}
\toprule
\textbf{Method} &
\textbf{Latency} &
\textbf{Price Diff.} &
\textbf{Order Value} \\
& \textbf{(s)} & \textbf{(\textperthousand)} & \textbf{Dev. (\textperthousand)} \\
\midrule

\method{XGBoost}        & 14.57 & 0.31 & 0.41 \\
\method{MLP}            & \underline{13.31} & 0.35 & 0.22 \\
\method{TabNet}         & 18.17 & 0.50 & 0.63 \\
\method{TCN}            & 16.89 & 0.33 & 0.22 \\
\method{LSTM}           & 27.98 & 0.40 & 0.30 \\
\method{TRA}            & 23.07 & 0.35 & 0.23 \\
\method{DoubleEnsemble} & 17.48 & 0.35 & 0.16 \\
\method{GATs}           & 17.45 & \underline{0.27} & \textbf{0.08} \\
\midrule

\method{MacroHFT}       & \textbf{5.32} & 0.27 & 0.14 \\
\midrule

\method{DeepSeek}       & 53.26  & 2.37 & 0.37 \\
\method{Qwen}           & 90.61  & \textbf{0.18} & 0.18 \\
\method{FinGPT}         & 416.89 & 1.88 & \underline{0.12} \\
\midrule

\method{FinMem}         & 545.33 & 2.19 & 1.94 \\
\method{FinAgent}       & 500.56 & 1.38 & 0.26 \\\bottomrule
\end{tabular}}
\vspace{-8pt}
\end{wraptable}

\stitle{Paper execution discrepancies.}
To investigate a possible source of the performance gap between backtesting and paper trading, we examine how intended trading decisions differ from their actual executions in Table~\ref{tab:papertrading_execution_gap}. We measure three discrepancies: \emph{latency}, the time in seconds between generating a trading decision and its execution; \emph{price difference}, the relative difference between the intended and executed prices; and \emph{order value deviation}, the difference between the intended and executed order values normalized by the current account capital.
We first observe that execution discrepancies differ markedly across methodological families. ML methods have an average latency of $18.6$ seconds and an average price difference of $0.36\text{\textperthousand}$, whereas these values increase to $186.9$ seconds and $1.48\text{\textperthousand}$ for LLM-based methods, and to $522.9$ seconds and $1.79\text{\textperthousand}$ for trading agents. Thus, methods that appear comparable under offline backtesting can encounter substantially different execution conditions once deployed in an exchange environment.
Second, however, larger execution discrepancies do not directly translate into larger backtest-to-paper-trading performance gaps. The absolute cumulative-return gap shows little Pearson correlation coefficient ($r$) \citep{benesty2009pearson} with latency ($r\approx0.05$), price difference ($r\approx-0.15$), or order value deviation ($r\approx0.19$). This suggests that execution effects are more complex than a single latency, price, or order-size penalty and cannot by themselves explain the observed variation in prospective performance.

\subsection{Live Trading}
\textbf{RQ3:} \textit{How closely do backtesting and paper trading reflect live trading performance?}

\begin{table*}[t]
\centering
\scriptsize
\setlength{\tabcolsep}{3pt}
\caption{Backtest (BT), paper trading (PT) and live trading (LT) performance.}
\label{tab:live}
\resizebox{\linewidth}{!}{
\begin{tabular}{l*{15}{c}}
\toprule
\multirow{2}{*}{\textbf{Method}} &
\multicolumn{3}{c}{\textbf{Cum. Ret. (\%)$\uparrow$}} &
\multicolumn{3}{c}{\textbf{Alpha (\%)$\uparrow$}} &
\multicolumn{3}{c}{\textbf{Sharpe$\uparrow$}} &
\multicolumn{3}{c}{\textbf{Max DD (\%)$\downarrow$}} &
\multicolumn{3}{c}{\textbf{Vol. (\%)$\downarrow$}} \\
\cmidrule(lr){2-4}
\cmidrule(lr){5-7}
\cmidrule(lr){8-10}
\cmidrule(lr){11-13}
\cmidrule(lr){14-16}
& \textbf{BT} & \textbf{PT} & \textbf{LT}
& \textbf{BT} & \textbf{PT} & \textbf{LT}
& \textbf{BT} & \textbf{PT} & \textbf{LT}
& \textbf{BT} & \textbf{PT} & \textbf{LT}
& \textbf{BT} & \textbf{PT} & \textbf{LT} \\
\midrule

\method{Buy\&Hold}
& \multicolumn{3}{c}{\textbf{11.63}}
& \textbf{0.00} & \underline{0.00} & \textbf{0.00}
& \textbf{1.59} & \underline{1.59} & \textbf{1.59}
& \multicolumn{3}{c}{\underline{7.48}}
& \multicolumn{3}{c}{40.66} \\

\midrule

\method{LSTM}
& -2.25 & -0.48 & -0.33
& -54.89 & -107.36 & -90.88
& -1.96 & -5.53 & -4.88
& 8.12 & 9.29 & 7.66
& \textbf{24.86} & \textbf{18.75} & \textbf{17.25} \\

\method{TRA}
& -9.21 & -10.82 & -14.74
& -160.46 & -235.95 & -284.03
& -2.18 & -3.11 & -4.41
& 18.38 & 19.14 & 21.49
& 58.71 & 58.26 & 51.85 \\

\method{XGBoost}
& -3.60 & 0.34 & -0.79
& -9.51 & -52.32 & -57.90
& -0.11 & -1.31 & -1.59
& 10.24 & 8.73 & 9.00
& \underline{28.54} & \underline{27.74} & \underline{25.20} \\

\midrule

\method{Qwen}
& \underline{6.18} & \underline{10.10} & \underline{3.72}
& \underline{-0.63} & \textbf{15.80} & \underline{-28.10}
& \underline{1.34} & \textbf{1.85} & \underline{0.34}
& \textbf{4.85} & \textbf{7.27} & \textbf{6.06}
& 29.91 & 39.28 & 32.02 \\

\bottomrule
\end{tabular}}
\end{table*}

\begin{figure}[t]
    \centering
    \includegraphics[width=\linewidth]{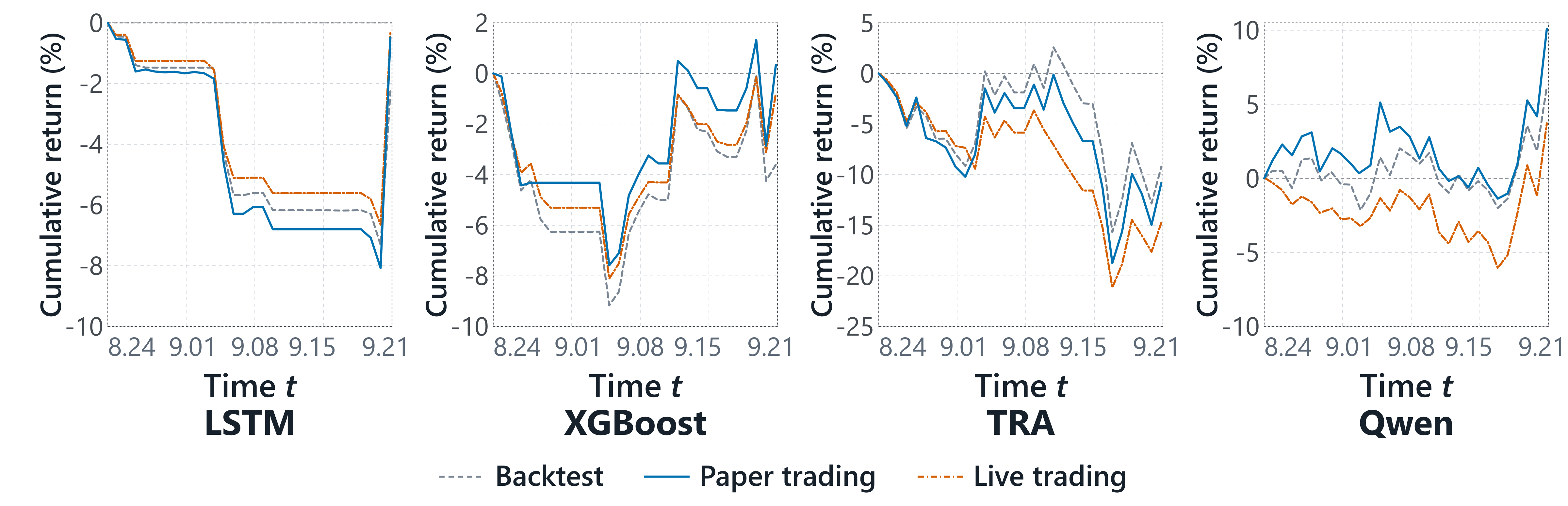}
    \vspace{-8mm}
    \caption{Cumulative returns of backtesting, paper trading and live trading.}
    \label{fig:bt-pt-lt}
\end{figure}

\stitle{Live trading performance.}
We report the cumulative-return curves of live trading from 2026-08-24 to 2026-09-21 in Fig.~\ref{fig:live-trading}, and summarize the corresponding performance in Table~\ref{tab:live}.
We first observe limited profitability under real-money trading. Only one of the four methods is profitable, and none outperforms the Bitcoin \method{Buy\&Hold}. The median return across the four methods is $-0.56\%$, despite the market gaining $11.63\%$ over the same period.
Second, even the strongest live-trading method fails to generate positive excess performance relative to the market. All four methods exhibit negative alpha, and only \method{Qwen} maintains a positive Sharpe ratio. Thus, methods selected through the preceding evaluation stages can still fail to deliver positive risk-adjusted excess performance once deployed with real capital.

\begin{wraptable}{r}{0.50\columnwidth}
\vspace{-8pt}
\centering
\scriptsize
\caption{Live trading execution discrepancies.}
\label{tab:livetrading_execution_gap}
\setlength{\tabcolsep}{3pt}

\resizebox{0.50\columnwidth}{!}{
\begin{tabular}{lccc}
\toprule
\textbf{Method} &
\textbf{Latency} &
\textbf{Price Diff.} &
\textbf{Order Value} \\
&
\textbf{(s)} &
\textbf{(\textperthousand)} &
\textbf{ Dev. (\textperthousand)} \\
\midrule

\method{LSTM}
& \textbf{10.63} & 2.63 & 1.23 \\

\method{TRA}
& 14.23 & 2.28 & 0.50 \\

\method{XGBoost}
& \underline{13.57} & \underline{1.39} & \underline{0.41} \\

\midrule

\method{Qwen}
& 151.00 & \textbf{1.13} & \textbf{0.27} \\

\bottomrule
\end{tabular}}
\vspace{-6pt}
\end{wraptable}

\stitle{Backtest, paper and live trading comparison.}
We further compare backtesting, paper trading, and live trading from 2026-08-24 to 2026-09-21. Table~\ref{tab:live} reports performance under all three settings, while Fig.~\ref{fig:bt-pt-lt} compares their cumulative-return curves.
We first observe that moving from paper trading to live trading introduces an additional performance gap. Three of the four methods achieve lower cumulative returns in live trading, with the paper-to-live return gap reaching $6.38$ pp. Thus, exchange-based paper trading does not fully reproduce real-money outcomes even when the market period and method configurations are held fixed.
Second, paper trading nevertheless provides a closer approximation to live trading than backtesting overall. Averaged across the four methods, the absolute paper-to-live discrepancy is smaller than the backtest-to-live discrepancy for all five evaluation metrics. The improvement is particularly pronounced for risk-adjusted performance: the average absolute gap decreases from $58.85$ pp to $28.51$ pp for alpha and from $1.91$ to $0.94$ for the Sharpe ratio, reductions of approximately $52\%$ and $51\%$, respectively. In comparison, the reductions are about $9\%$ for cumulative return and maximum drawdown and $11\%$ for volatility. These results show that paper trading substantially narrows some dimensions of the gap to real-money trading, but does not eliminate the need for direct live evaluation.

\stitle{Live execution discrepancies.}
To investigate a possible source of the remaining performance gap between paper trading and live trading, we examine execution discrepancies under real-money trading in Table~\ref{tab:livetrading_execution_gap}. We use the same three measures as in paper trading: latency, price difference, and order value deviation.
We first observe that execution discrepancies become substantially larger when moving from paper trading to real-money trading. For the four deployed methods, the average price difference increases from approximately $0.31\text{\textperthousand}$ in paper trading to $1.86\text{\textperthousand}$ in live trading, an increase of nearly $6\times$. The average order value deviation also increases from $0.28\text{\textperthousand}$ to $0.60\text{\textperthousand}$, more than doubling. In contrast, average execution latency changes much less, from approximately $39$ to $47$ seconds.
Second, the increase is therefore concentrated more strongly in how trading decisions are realized than in how long they take to execute. In particular, the much larger deviations in execution price and order value provide a plausible source of the remaining paper-to-live performance gap. Paper trading captures exchange-based execution, but real-money trading introduces additional discrepancies between intended and realized trades that are not fully reproduced in the simulated environment.

\subsection{Discussion of Insights and Future Directions}\label{sec.finding}
\stitle{Backtest success does not reliably transfer to prospective trading outcomes.}
Strong performance in historical backtesting can change substantially once methods are evaluated prospectively. The methodological pattern observed in the 2025 backtest shifts markedly in paper trading, where methods that were previously weak become competitive and many strong ML methods lose their advantage. More importantly, this discrepancy cannot be attributed only to changing market conditions: even over the same market period, backtesting and paper trading can produce materially different profitability and risk profiles. These findings show that historical backtesting alone is insufficient for estimating prospective trading performance. \emph{Future work} should therefore emphasize robustness across both changing market conditions and execution environments rather than optimizing a single historical backtest.

\stitle{Paper trading narrows but does not close the gap to real-money trading.}
Paper trading provides a meaningful intermediate stage between backtesting and live deployment. Across the evaluated metrics, paper-trading results are consistently closer on average to live-trading outcomes than backtesting results are, with particularly large improvements in alpha and Sharpe ratio. However, measurable discrepancies remain between paper and live trading when the same methods are evaluated over the same market period. \textit{Future work} should thus adopt a staged evaluation protocol in which paper trading follows backtesting to assess prospective performance, providing a necessary intermediate stage without the financial risks of live trading before final live validation and deployment.

\stitle{Current trading methods do not sustain market-beating performance under live deployment.}
A more fundamental limitation emerges under real-money trading: none of the deployed methods outperforms the market, and all exhibit negative alpha. This suggests that the main challenge is not merely identifying the strongest method within a benchmark, but developing strategies that can sustain excess returns once exposed to changing markets and real execution. \emph{Future work} should therefore move beyond optimizing absolute backtest returns and explicitly target persistent excess returns, deployment robustness, and consistency between offline evaluation and realized trading outcomes.

\section{Conclusion}
We introduced a unified benchmark for evaluating AI trading methods beyond historical backtesting, covering representative ML, RL, LLM-based, and agent-based approaches under a three-stage protocol spanning historical backtesting, exchange-based paper trading, and real-money live trading. By progressively increasing both temporal and execution realism, the benchmark enables systematic measurement of the backtest-to-realization gap. Our results show that strong historical performance does not reliably transfer to prospective trading, while paper trading provides a closer approximation to live performance than backtesting but still leaves a meaningful gap under real-money execution. Moreover, none of the evaluated methods sustains market-beating performance in live trading, highlighting the need to move beyond optimizing performance in a single historical setting. Finally, we release an open-source system and public platform that support reproducible evaluation across all three stages while continuously updating paper-trading and live-trading results on newly arriving market data.

\clearpage
\newpage
\section*{AI Use Statement}
We used AI tools solely to polish the language and improve the readability of the manuscript. These tools were not used to develop research ideas, design or implement methods, conduct experiments, or analyze and interpret results. The authors take full responsibility for the final content of this paper.

\section*{Ethics Statement}
We have not identified specific ethical concerns arising from this study. The benchmark is intended for research and evaluation purposes, and the reported results should not be interpreted as investment advice or guarantees of future trading performance.

\section*{Reproducibility Statement}
Implementation details are provided in the appendix to support reproduction of our experiments. We have also made the source code publicly available to facilitate independent verification and further research.

\bibliography{iclr2027_conference}
\bibliographystyle{iclr2027_conference}

\clearpage
\newpage
\appendix
\counterwithin{figure}{section}

\section{Evaluated Methods}\label{app.methods}

We provide detailed descriptions of the methods evaluated in our benchmark. The benchmark covers four major paradigms of AI-based trading---machine learning, reinforcement learning, LLM-based trading, and trading agents---as well as a passive Buy\&Hold baseline. The methods differ substantially in how they represent market information and generate trading decisions, ranging from
numerical prediction models to autonomous LLM-based trading systems. We summarize the evaluated methods in Table~\ref{tab:method_overview}. The table compares the information used by each method, including direct market data, engineered market factors, and textual news information, together with key aspects of its trading formulation, including whether it dynamically selects assets, its permitted trading direction, and whether it explicitly generates risk-control instructions such as stop-loss or target prices.

\begin{table*}[tbp]
    \centering
    \footnotesize
    \caption{Overview of the evaluated trading methods.}
    \label{tab:method_overview}
    \addtolength{\tabcolsep}{0.7mm}
    \resizebox{\textwidth}{!}{%
    \begin{tabular}{@{}llcccccc@{}}
    \toprule
    \multirow{2}{*}{Method}
    & \multirow{2}{*}{Category}
    & \multicolumn{3}{c}{Information}
    & \multicolumn{3}{c}{Trading Decision} \\
    \cmidrule(lr){3-5}
    \cmidrule(lr){6-8}
    &
    & Market Data
    & Factors
    & News
    & Asset Sel.
    & Direction
    & Risk Ctrl. \\
    \midrule

    \method{Buy\&Hold}
    & Passive
    & $\times$
    & $\times$
    & $\times$
    & $\times$
    & Long-only
    & $\times$ \\

    \midrule

    \method{LightGBM}
    & ML
    & $\times$
    & $\checkmark$
    & $\times$
    & $\checkmark$
    & Long-only
    & $\times$ \\

    \method{CatBoost}
    & ML
    & $\times$
    & $\checkmark$
    & $\times$
    & $\checkmark$
    & Long-only
    & $\times$ \\

    \method{Linear}
    & ML
    & $\times$
    & $\checkmark$
    & $\times$
    & $\checkmark$
    & Long-only
    & $\times$ \\

    \method{XGBoost}
    & ML
    & $\times$
    & $\checkmark$
    & $\times$
    & $\checkmark$
    & Long-only
    & $\times$ \\

    \method{MLP}
    & ML
    & $\times$
    & $\checkmark$
    & $\times$
    & $\checkmark$
    & Long-only
    & $\times$ \\

    \method{TabNet}
    & ML
    & $\times$
    & $\checkmark$
    & $\times$
    & $\checkmark$
    & Long-only
    & $\times$ \\

    \method{TCN}
    & ML
    & $\times$
    & $\checkmark$
    & $\times$
    & $\checkmark$
    & Long-only
    & $\times$ \\

    \method{GRU}
    & ML
    & $\times$
    & $\checkmark$
    & $\times$
    & $\checkmark$
    & Long-only
    & $\times$ \\

    \method{LSTM}
    & ML
    & $\times$
    & $\checkmark$
    & $\times$
    & $\checkmark$
    & Long-only
    & $\times$ \\

    \method{SFM}
    & ML
    & $\times$
    & $\checkmark$
    & $\times$
    & $\checkmark$
    & Long-only
    & $\times$ \\

    \method{GeneralPtNN}
    & ML
    & $\times$
    & $\checkmark$
    & $\times$
    & $\checkmark$
    & Long-only
    & $\times$ \\

    \method{ALSTM}
    & ML
    & $\times$
    & $\checkmark$
    & $\times$
    & $\checkmark$
    & Long-only
    & $\times$ \\

    \method{TFT}
    & ML
    & $\times$
    & $\checkmark$
    & $\times$
    & $\checkmark$
    & Long-only
    & $\times$ \\

    \method{Transformer}
    & ML
    & $\times$
    & $\checkmark$
    & $\times$
    & $\checkmark$
    & Long-only
    & $\times$ \\

    \method{KRNN}
    & ML
    & $\times$
    & $\checkmark$
    & $\times$
    & $\checkmark$
    & Long-only
    & $\times$ \\

    \method{Localformer}
    & ML
    & $\times$
    & $\checkmark$
    & $\times$
    & $\checkmark$
    & Long-only
    & $\times$ \\

    \method{TRA}
    & ML
    & $\times$
    & $\checkmark$
    & $\times$
    & $\checkmark$
    & Long-only
    & $\times$ \\

    \method{AdaRNN}
    & ML
    & $\times$
    & $\checkmark$
    & $\times$
    & $\checkmark$
    & Long-only
    & $\times$ \\

    \method{TCTS}
    & ML
    & $\times$
    & $\checkmark$
    & $\times$
    & $\checkmark$
    & Long-only
    & $\times$ \\

    \method{ADD}
    & ML
    & $\times$
    & $\checkmark$
    & $\times$
    & $\checkmark$
    & Long-only
    & $\times$ \\

    \method{DoubleEnsemble}
    & ML
    & $\times$
    & $\checkmark$
    & $\times$
    & $\checkmark$
    & Long-only
    & $\times$ \\

    \method{GATs}
    & ML
    & $\times$
    & $\checkmark$
    & $\times$
    & $\checkmark$
    & Long-only
    & $\times$ \\

    \method{Sandwich}
    & ML
    & $\times$
    & $\checkmark$
    & $\times$
    & $\checkmark$
    & Long-only
    & $\times$ \\

    \method{IGMTF}
    & ML
    & $\times$
    & $\checkmark$
    & $\times$
    & $\checkmark$
    & Long-only
    & $\times$ \\

    \midrule

    \method{MacroHFT}
    & RL
    & $\times$
    & $\checkmark$
    & $\times$
    & $\times$
    & Long-only
    & $\times$ \\

    \method{EarnHFT}
    & RL
    & $\times$
    & $\checkmark$
    & $\times$
    & $\times$
    & Long-only
    & $\times$ \\

    \midrule

    \method{DeepSeek}
    & LLM
    & $\checkmark$
    & $\times$
    & $\times$
    & $\checkmark$
    & Long/Short
    & $\checkmark$ \\

    \method{Qwen}
    & LLM
    & $\checkmark$
    & $\times$
    & $\times$
    & $\checkmark$
    & Long/Short
    & $\checkmark$ \\

    \method{FinGPT}
    & LLM
    & $\times$
    & $\times$
    & $\checkmark$
    & $\checkmark$
    & Long-only
    & $\times$ \\

    \midrule

    \method{FinMem}
    & Agent
    & $\checkmark$
    & $\times$
    & $\checkmark$
    & $\checkmark$
    & Long/Short
    & $\times$ \\

    \method{FinAgent}
    & Agent
    & $\checkmark$
    & $\times$
    & $\checkmark$
    & $\checkmark$
    & Long/Short
    & $\times$ \\

    \method{TradingAgents}
    & Agent
    & $\checkmark$
    & $\times$
    & $\checkmark$
    & $\checkmark$
    & Long/Short
    & $\times$ \\

    \bottomrule
    \end{tabular}}
\end{table*}

\stitle{Machine learning methods.}
We evaluate a broad collection of machine learning methods that generate trading signals by predicting future market movements from numerical market observations. Concretely, we include:

\begin{itemize}

    \item \method{Linear} \citep{yang2020qlib} learns a linear mapping from market features to the prediction target, providing a simple baseline for assessing whether more complex nonlinear models offer additional predictive value.

    \item \method{LightGBM} \citep{ke2017lightgbm} employs gradient-boosted decision trees with leaf-wise tree growth and efficient feature discretization to model nonlinear relationships between market features and future returns.

    \item \method{XGBoost} \citep{chen2016xgboost} uses regularized gradient-boosted decision trees to iteratively correct prediction errors while controlling model complexity.

    \item \method{CatBoost} \citep{prokhorenkova2018catboost} is a gradient-boosting framework that improves robustness through ordered boosting and regularization strategies, and is used here as another tree-based predictor for market signals.

    \item \method{MLP} \citep{yang2020qlib} applies a multi-layer feed-forward neural network to learn nonlinear mappings from numerical market features to future market movements.

    \item \method{TabNet} \citep{arik2021tabnet} uses sequential attention to select and transform informative subsets of input features at different decision steps, providing an attention-based neural model for structured market data.

    \item \method{TCN} \citep{bai2018empirical} models historical market sequences using causal and dilated temporal convolutions, allowing the model to capture long-range temporal dependencies without recurrent computation.

    \item \method{GRU} \citep{cho2014learning} employs gated recurrent units to summarize historical market observations while controlling how past information is retained and updated over time.

    \item \method{LSTM} \citep{hochreiter1997long} uses gated recurrent memory cells to capture
    long-term dependencies in financial time series and predict future market
    movements from historical observations.

    \item \method{SFM} \citep{zhang2017stock} introduces state-frequency memory to decompose temporal
    dynamics into multiple frequency components, enabling the model to capture
    trading patterns that occur at different frequencies.

    \item \method{GeneralPtNN} \citep{yang2020qlib} provides a general PyTorch-based prediction
    framework in which different neural backbones can be trained through a
    unified interface. We use the model configuration specified in our
    implementation to learn predictive signals from historical market
    features.

    \item \method{ALSTM} \citep{qin2017dual} augments an LSTM-based temporal encoder with an
    attention mechanism that assigns different importance to historical
    observations when producing the final market prediction.

    \item \method{TFT} \citep{lim2021temporal} combines recurrent sequence modeling, gating and
    variable-selection mechanisms, and temporal attention to capture both
    local and long-range dependencies in multivariate time series.

    \item \method{Transformer} \citep{vaswani2017attention} relies on self-attention to directly model
    dependencies among observations at different time steps, providing a
    non-recurrent architecture for learning temporal market patterns.

    \item \method{KRNN} \citep{yang2020qlib} employs multiple recurrent branches to capture
    heterogeneous temporal patterns in market sequences and combines their
    representations for prediction.

    \item \method{Localformer} \citep{yang2020qlib} extends Transformer-style sequence modeling
    with local convolutional operations, combining local temporal pattern
    extraction with attention-based modeling of longer-range dependencies.

    \item \method{TRA} \citep{lin2021tra} introduces a temporal routing adaptor with multiple
    predictors for different trading patterns and a router that dynamically
    assigns samples to these predictors. Optimal transport is used to guide
    the discovery and assignment of latent market patterns.

    \item \method{AdaRNN} \citep{du2021adarnn} addresses temporal distribution shifts by learning
    transferable representations across different time periods, reducing the
    discrepancy between historical market regimes during sequence
    forecasting.

    \item \method{TCTS} \citep{wu2021tcts} exploits temporally correlated auxiliary prediction
    tasks and dynamically schedules them during training so that information
    from related forecasting horizons can improve the target prediction task.

    \item \method{ADD} \citep{tang2020add} uses augmented disentanglement distillation to separate
    market-wide and excess-return information from noisy financial features,
    together with self-distillation and data augmentation to improve stock
    trend prediction.

    \item \method{DoubleEnsemble} \citep{zhang2020doubleensemble} combines learning-trajectory-based sample
    reweighting with shuffling-based feature selection, aiming to reduce
    overfitting to noisy samples and irrelevant features in non-stationary
    financial data.

    \item \method{GATs} \citep{velickovic2018graph} applies graph attention to model relations among
    financial entities, allowing the contribution of related entities to be
    weighted adaptively when constructing representations for prediction.

    \item \method{Sandwich} \citep{yang2020qlib} stacks convolutional and recurrent encoding
    modules to jointly capture local temporal patterns and longer-range
    sequential dependencies in market observations.

    \item \method{IGMTF} \citep{xu2021igmtf} constructs instance-wise graphs over multivariate
    time-series observations and aggregates information across related
    variables and historical instances, explicitly modeling dependencies that
    extend across both variables and time.

\end{itemize}

\stitle{Reinforcement learning methods.}
Unlike predictive ML models, reinforcement learning methods directly optimize
trading policies according to rewards generated through sequential interaction
with the market. We evaluate:

\begin{itemize}

    \item \method{EarnHFT}~\citep{qin2024earnhft}  develops a hierarchical reinforcement learning
    framework for high-frequency trading. It trains multiple specialized
    low-level trading agents and uses a higher-level routing policy to select
    appropriate agents under different market conditions, reducing the
    difficulty of learning over extremely long high-frequency trajectories.

    \item \method{MacroHFT}~\citep{zong2024macrohft}  further incorporates market context and memory
    into hierarchical high-frequency trading. It trains context-conditioned
    sub-agents for different market regimes and uses a memory-augmented
    hyper-agent to combine their decisions and adapt the final trading policy
    to changing market conditions.

\end{itemize}

\stitle{LLM-based trading methods.}
We further evaluate approaches that use large language models to interpret
market information and directly generate trading decisions. These methods
differ from conventional numerical predictors in that the trading decision is
produced through language-model reasoning over structured or textual market
context.

\begin{itemize}

    \item \method{DeepSeek} \citep{guo2025deepseekr1} serves as a general-purpose LLM trading baseline.
    We provide the model with the market context available at each decision
    step and prompt it to reason about the current market condition and
    generate the corresponding trading action.

    \item \method{Qwen} \citep{yang2025qwen3} is evaluated under the same direct LLM trading
    setting, where standardized market information is converted into the
    model's input context and the model generates a trading decision through
    language-based reasoning.

    \item \method{FinGPT} \citep{yang2023fingpt} is a finance-oriented open-source LLM framework
    developed with financial data curation and lightweight adaptation. In the
    trading setting, it leverages financial-domain representations and
    language reasoning to transform market information into trading signals.

\end{itemize}

\stitle{Trading agents.}
Trading agents extend direct LLM-based decision making by introducing explicit
agent architectures for organizing information, maintaining experience, or
coordinating multiple reasoning roles. We evaluate:

\begin{itemize}

    \item \method{FinMem} \citep{li2025investorbench} equips an LLM trading agent with a layered memory
    architecture that organizes financial information over different temporal
    horizons. Its profiling, memory, and decision-making modules allow the
    agent to retrieve relevant historical experience and use it to inform
    subsequent investment decisions.

    \item \method{FinAgent} \citep{zhang2024finagent} is a multimodal trading agent that integrates
    numerical market data, financial text, and visual market information.
    It combines tool-augmented market analysis with diversified memory and
    dual-level reflection to learn from historical experience and improve
    subsequent trading decisions.

    \item \method{TradingAgents} \citep{xiao2024tradingagents} models a trading firm as a collection of
    specialized LLM agents. Different agents perform technical, fundamental,
    and sentiment analysis, conduct bullish and bearish debates, assess risk,
    and synthesize these views into a final trading decision through structured
    multi-agent collaboration.

\end{itemize}

\stitle{Passive baseline.}
We additionally include \method{Buy\&Hold}, which purchases the target asset
at the beginning of the evaluation period and maintains the position throughout
the period without active trading. It provides a passive market benchmark for
measuring whether an active trading method generates excess return.

\section{Implementation Details}\label{app.hyperparams}
In this section, we elaborate on the full implementation details and hyperparameter configurations introduced in Section~\ref{sec.setting}.

\subsection{General settings}\label{app.general-setting}
\noindent\textbf{Optimizer.} For all experiments, we use the Adam optimizer~\citep{kingma2015adam}.

\noindent\textbf{Environment.} The experimental server has the following configuration:
\begin{itemize}
    \item Operating system: AlmaLinux 9.8 
    \item CPU: AMD EPYC 7513
    \item Memory: 188.2 GiB RAM 
    \item GPU: 1$\times$ NVIDIA L40, with 46,068 MB memory
    \item NVIDIA driver: 610.43.02
    \item CUDA Toolkit: 12.6.85
    \item PyTorch: 2.12.0 
    \item Python: 3.10.20 
\end{itemize}

\subsection{Details of the assets.}\label{app.assets}
We construct the tradable universe using 10 large-cap and widely traded cryptocurrencies. Ordered by market capitalization at the time of benchmark construction, these assets are Bitcoin (BTC), Ethereum (ETH), XRP (XRP), Dogecoin (DOGE), Cardano (ADA), TRON (TRX), Chainlink (LINK), Litecoin (LTC), Hedera (HBAR), and OKB (OKB). The selected assets cover different types of cryptocurrencies, including payment-oriented assets, native assets of smart-contract platforms, and ecosystem or utility tokens. 
These 10 assets define the candidate trading universe of the benchmark. However, individual methods do not necessarily trade all assets. For example, the ML methods perform cross-asset prediction and select assets from this universe, whereas the RL methods operate on a predefined asset, ETH, as described in Section~\ref{sec.setting}.

\subsection{Hyperparameter settings.}
We use the official implementations for the ML baselines whenever available.
Unless otherwise specified, neural models are trained with Adam using an initial
learning rate of $10^{-3}$, a batch size of 2,000, at most 200 epochs, and an
early-stopping patience of 20. The random seed is set to 42 where supported.
For sequence-based models, we use a 60-bar lookback window. We main test 5 minutes, 15 minutes, 1 hour and 4 hours, and finally choose the best time frame for simulate trading.

For Linear \citep{yang2020qlib}, we use ordinary least squares without an intercept or additional regularization.

For LightGBM \citep{ke2017lightgbm}, XGBoost \citep{chen2016xgboost}, and CatBoost \citep{prokhorenkova2018catboost}, we use up to 1,000 boosting rounds with an early-stopping patience of 50. Model-specific tree parameters, such as the learning rate and tree depth, follow the corresponding implementation defaults.

For DoubleEnsemble \citep{zhang2020doubleensemble}, we use six LightGBM-based component models with 100 boosting
rounds per component, while enabling both sample reweighting and feature selection.

For MLP \citep{yang2020qlib}, we use one hidden layer with 256 units and a dropout rate of 0.05.
Different from the other neural baselines, MLP is optimized using SGD for up to
300 update steps, with an early-stopping patience of 50 evaluation rounds.

For LSTM \citep{hochreiter1997long}, GRU \citep{cho2014learning}, and ALSTM \citep{qin2017dual}, we use two recurrent layers with a hidden dimension of
64 and no dropout.

For GeneralPTNN \citep{yang2020qlib}, we use a two-layer GRU backbone with a hidden dimension of 64.

For GATs \citep{velickovic2018graph}, HIST \citep{xu2021hist}, and IGMTF \citep{xu2021igmtf}, we use a GRU backbone with two layers and a hidden
dimension of 64. These models retain their native cross-sectional batching
mechanisms.

For SFM \citep{zhang2017stock}, we set the hidden dimension to 64 and the frequency dimension to 10,
and optimize the model using SGD.

For TCN \citep{bai2018empirical}, we use two temporal convolutional levels with 128 channels and a kernel
size of 5.

For KRNN \citep{yang2020qlib}, we set both the CNN and RNN hidden dimensions to 64, use a convolution
kernel size of 3, and employ three parallel recurrent branches.

For Sandwich \citep{yang2020qlib}, the two CNN stages have hidden dimensions of 64 and 32, while the
two recurrent stages use hidden dimensions of 16 and 8, respectively. Three
parallel recurrent branches are used.

For AdaRNN \citep{du2021adarnn}, we use two domains and 40 pretraining epochs, with the transfer-loss
weight set to 0.5.

For ADD \citep{tang2020add}, we use two recurrent layers with a hidden dimension of 64. The batch
size is increased to 5,000, with $\gamma=0.1$ and $\mu=0.05$.

For Transformer \citep{vaswani2017attention} and Localformer \citep{yang2020qlib}, we use a model dimension of 64, two attention
heads, and two layers. The learning rate is reduced to $10^{-4}$ with a weight
decay of $10^{-3}$. We train for up to 100 epochs with an early-stopping patience
of 5. The standard and time-series implementations use batch sizes of 2,048 and
8,192, respectively.

For TabNet \citep{arik2021tabnet}, we use a representation dimension of 64, a relaxation factor of 1.3,
and a virtual batch size of 2,048. The learning rate is set to $10^{-2}$ with a
batch size of 4,096. The model is pretrained for 50 epochs before supervised
training for up to 100 epochs.

For TCTS \citep{wu2021tcts}, we use two layers with a hidden dimension of 64 and three optimization
steps. Both the forecasting and weighting networks use a learning rate of
$5\times10^{-7}$.

For TRA \citep{lin2021tra}, we use a two-layer GRU backbone with a hidden dimension of 64 and a
single latent state. The batch size is 256, and the model is trained for up to
500 epochs with an early-stopping patience of 50.

For TFT \citep{lim2021temporal}, we use a model dimension of 64, four attention heads, two layers, and a
dropout rate of 0.2. The learning rate is $10^{-3}$ with a weight decay of
$10^{-4}$ and a batch size of 128. The model is trained for up to 100 epochs
with an early-stopping patience of 10.



We use DeepSeek V4 Flash (\texttt{deepseek-v4-flash}) and Qwen 3.6 Max
Preview (\texttt{qwen3.6-max-preview}), both with thinking enabled. The two
baselines share the same prompt generator and executor so that only the
language model changes. At each decision, the prompt contains fifty 3-minute
bars summarized as price and technical-indicator sequences; fifty 4-hour bars
summarized as trend, volatility, and volume indicators; the funding rate and
open interest; account equity and current positions; and up to five recent
decisions. We do not supply news to either direct baseline. Each model returns
a JSON object containing \texttt{buy}, \texttt{sell}, or \texttt{hold}, a
proposed contract quantity, stop-loss, profit target, invalidation condition,
confidence, estimated risk, and justification. The executor supports
a separate close operation, but the current direct-model prompt does not
request it.

For FinGPT \citep{yang2023fingpt}, we use \texttt{NousResearch/Llama-2-13b-hf} as the base model and
initialize from the public \texttt{FinGPT/fingpt-sentiment\_llama2-13b\_lora}
adapter. The final deployed model is the news-focused Stage-2 adapter
\texttt{stage2\_news\_focused}. We perform two SFT stages. Stage 1 uses a
learning rate of $8\times10^{-5}$ and seed 42, while Stage 2 uses
$5\times10^{-5}$ and seed 43. Both stages are trained for one epoch with a
per-device batch size of 2 and 16 gradient-accumulation steps, corresponding to
an effective single-device batch size of 32. We use AdamW with weight decay
0.01, a linear learning-rate schedule, a warmup ratio of 0.03, and a maximum
sequence length of 512. Training is performed with 4-bit NF4 quantization and
FP16 computation.

The deployed LoRA adapter uses rank $r=8$, scaling factor
$\alpha=32$, and dropout 0.1, and is applied to the query, key, and value
projection modules. The training data combine
\texttt{DLT-Sentiment-News} and \texttt{stocktwits-crypto}. Stage 1 uses
60,000 samples from each source, whereas Stage 2 increases the proportion of
news data to 48,000 DLT samples and 12,000 StockTwits samples. A separate
validation set contains 7,200 DLT and 1,800 StockTwits samples, with balanced
sampling over negative, neutral, and positive sentiment classes.

For inference, we use label-logprob classification over the three sentiment
labels. Each label is scored by the mean token log-probability, and the
article-level sentiment score is defined as
$P(\mathrm{positive})-P(\mathrm{negative})$. The maximum input length is 512
tokens and the deployed inference batch size is 1. The model is loaded in FP16
without 4-bit or 8-bit inference quantization.

For trading-signal construction, parameters are selected separately for each
asset. The candidate strategies include asset-only, market-only, fixed
asset--market fusion, and adaptive fusion. The selected configurations use
asset-news top-$N$ values between 6 and 24, market-news top-$N$ values between
12 and 48, and signal thresholds between 0.03 and 0.15. Fixed-fusion weights
range from 0.65 to 0.80, while the adaptive configuration uses
$k=4$ with the asset weight clipped to $[0.65,0.95]$. Signals are evaluated
daily with a one-day lag. Each asset is assigned an independent capital bucket,
and the strategy operates in a long/flat regime.


For FinMem \citep{li2025investorbench}, we use \texttt{qwen3-max} as the decision model and Qdrant as the
vector-memory database. We retrieve the top five memories and use three-day
look-back and momentum windows. A 60-day warm-up period is used to initialize
the memory system. The initial importance scores for short-, mid-, long-term,
and reflection memories are set to 50, 60, 90, and 80, with recency decay
factors of 3, 90, 365, and 365 and importance decay factors of 0.92, 0.96,
0.96, and 0.98, respectively. Memory promotion thresholds are set to 55 and
85, while the reflection similarity threshold is 0.95. Qdrant stores
1,024-dimensional embeddings using cosine similarity. For each  daily decision, we
retain the current day and the preceding 10 days of context, with at most five
asset-specific and five market-wide news items.


For FinAgent, we use DeepSeek V4 Flash with thinking enabled for market-intelligence
summarization, low- and high-level reflection, and the final decision. In the
normal OKX data path, FinAgent receives up to 100 closed 4-hour open, high, low,
close, and volume bars, up to 30 days of news, the current account and signed
position, retrieved market-intelligence memories, price reflections, decision
reflections, and the current execution constraints. Memory retrieval uses a
default of five items, while market-intelligence retrieval returns at most
three items per query type. FinAgent returns XML fields \texttt{analysis},
\texttt{action}, and \texttt{reasoning}; \texttt{action} is one of
\texttt{BUY}, \texttt{SELL}, \texttt{HOLD}, or \texttt{CLOSE}. The model
does not output an executable contract quantity.

For TradingAgents, we use DeepSeek V4 Flash with thinking enabled for both quick- and deep-reasoning
roles. The deployed configuration selects the market and news input analysts.
Their reports are passed through one initial bull-research pass, the research
manager, the trader, one initial aggressive-risk pass, and the portfolio
manager. The bull--bear debate-round count and the three-way risk-discussion
round count are both set to zero, so the bear, conservative, and neutral
follow-up turns are skipped. The final model output contains one of five
ratings---\texttt{Buy}, \texttt{Overweight}, \texttt{Hold},
\texttt{Underweight}, or \texttt{Sell}---together with an executive summary,
investment thesis, and optional price target and time horizon.

\noindent\textbf{Historical direct-model replay.}
For provenance, the 2025 direct-model backtests reported in the main results
use BTC-USDT perpetual contracts, 4-hour decisions, 10$\times$ isolated
leverage, 10,000 USDT initial equity, and a 0.10\% transaction fee. The reported
\method{DeepSeek} trajectory is a resumed run whose early segment used the Pro
variant and later segment used the Flash variant; it should therefore be read
as a mixed-model trajectory rather than as a single fixed-model estimate.

\noindent\textbf{Prospective execution settings.}
In prospective paper trading, all four systems poll every 300 seconds but make
at most one decision for each newly closed UTC 4-hour bar. 
Perpetual positions use isolated
margin with fixed 2$\times$ leverage. The executor caps the absolute notional of
one asset at 15\% of current account equity and total gross notional across the
four assets at 60\%. The direct baselines skip order changes
below 10 USDT; FinAgent instead follows the exchange contract's minimum size
and lot-size rules.

The model output and the executable order are deliberately separated. For
\method{DeepSeek} and \method{Qwen}, the proposed quantity is reduced when it
exceeds either exposure cap. For \method{FinAgent}, \texttt{BUY} and
\texttt{SELL} target the largest long or short position allowed by the two
caps, \texttt{CLOSE} targets zero, and \texttt{HOLD} preserves the current
position. For \method{TradingAgents}, \texttt{Buy} and \texttt{Sell} map to
$+100\%$ and $-100\%$ of the per-asset cap, while \texttt{Overweight} and
\texttt{Underweight} map to $+75\%$ and $-75\%$ of that cap. The executor then
subtracts the current signed position, converts the notional difference to
contract units using the exchange contract value, and rounds down to the
permitted lot size. Prompt-proposed stop-loss, profit-target, and
natural-language sizing suggestions are recorded for analysis but do not
override these deterministic exposure limits or create exchange-native
conditional orders.


\subsection{Prompt design}
We report the prompt contracts used by the LLMs based methods. 

\begin{promptbox}{System Prompt for DeepSeek and Qwen}
\small
You are a professional cryptocurrency trading assistant. Produce risk-aware
4-hour decisions that preserve capital first and grow it second. Do not trade
merely to remain active; return \texttt{hold} when trend, momentum, volatility,
or multi-timeframe evidence is mixed, late, or noisy. Treat current account
value as principal. For every non-hold trade, the expected loss at the proposed
stop should normally be 0.5--2.0\% of principal and must not exceed 3.0\%.
Every non-hold decision must include a stop-loss, profit target, and explicit
invalidation condition. Return only valid JSON with the following fields for
each asset: \texttt{signal}, \texttt{quantity}, \texttt{stop\_loss},
\texttt{profit\_target}, \texttt{invalidation\_condition},
\texttt{justification}, \texttt{confidence}, and
\texttt{risk\_usd}. For \texttt{hold}, set \texttt{quantity} and
\texttt{risk\_usd} to zero. The percentage range and ceiling are model-side
sizing guidance recorded with the response; the executor does not enforce
\texttt{risk\_usd} or submit the proposed stop as a conditional order.
\end{promptbox}

\begin{promptbox}{Decision-Time Prompt for DeepSeek and Qwen}
\small
Current time: \texttt{\{timestamp\}}. Invocation count:
\texttt{\{decision\_index\}}. All time series are ordered from oldest to
newest. For each asset, the prompt supplies: current price; the latest ten
3-minute closes; 20-period exponential moving average, moving-average
convergence divergence, and 7- and 14-period relative-strength-index series
computed from the latest fifty 3-minute bars; 20- and 50-period exponential
moving averages, 3- and 14-period average true range, current and mean volume,
and the latest ten moving-average-convergence-divergence and 14-period
relative-strength-index values computed from fifty 4-hour bars; funding rate;
open interest; available cash; account value; and current perpetual positions.
The prompt then appends up to five recent decisions, if available. Using only
this market and account state, return the JSON decision required by the system
prompt. No news is supplied to these two baselines.
\end{promptbox}

\begin{promptbox}{System and Decision Prompt for FinAgent}
\small
\textbf{System role.} You are an expert trader making disciplined 4-hour
cryptocurrency decisions from market information, news summaries, account
state, retrieved memories, and reflections. The execution context states
whether the instrument is spot or a perpetual contract. In perpetual mode,
\texttt{BUY} targets a bounded long position, \texttt{SELL} targets a bounded
short position, \texttt{CLOSE} targets a flat position, and \texttt{HOLD}
preserves the current signed position. Do not infer quantity, or
account capability beyond the supplied execution constraints.

\textbf{Decision instruction.} Instrument: \texttt{\{asset\}}; date:
\texttt{\{date\}}; reference price: \texttt{\{price\}}; cash:
\texttt{\{cash\}}; current position: \texttt{\{position\}}; cumulative and
recent returns: \texttt{\{returns\}}. Consider the latest and retrieved past
market-intelligence summaries, short-, medium-, and long-horizon price
reflections, reflections on prior trading decisions, and the supplied
execution constraints. Assess whether each item is positive, negative, neutral,
stale, or unrelated; do not chase a missed entry or panic-sell a missed exit.
Return only XML with \texttt{analysis}, one \texttt{action} from
\texttt{BUY}/\texttt{SELL}/\texttt{HOLD}/\texttt{CLOSE}, and
\texttt{reasoning}. The model supplies direction and reasoning; the executor
computes the contract quantity.
\end{promptbox}

The current FinAgent template was inherited from a broader multimodal design,
but the cryptocurrency adapter used here fills the OHLCV, news, memory,
reflection, account, and execution-context fields. Optional guidance,
sentiment, economics, and visual sections that are unavailable in a run are
left empty rather than inferred. One inherited instruction fragment still
enumerates only \texttt{BUY}, \texttt{SELL}, and \texttt{HOLD}; the task
description and output schema explicitly admit \texttt{CLOSE}, and both the
response parser and executor accept it. Those components are authoritative for
the deployed system despite the residual three-action fragment.

\begin{promptbox}{Role Prompts for TradingAgents}
\small
\textbf{Market analyst.} Analyze the current cryptocurrency pair using local
OHLCV data and at most eight complementary indicators. Emphasize trend,
momentum, volatility regime, volume confirmation, support and resistance,
trend exhaustion, and downside risk; return a detailed evidence-grounded market
report.

\textbf{News analyst.} Analyze recent asset-specific and global news, focusing
on macro liquidity, rates, dollar strength, regulation, fund-flow narratives,
exchange events, security incidents, protocol upgrades, and broad risk
appetite; return an evidence-grounded news report.

\textbf{Trader.} Given the research plan and analyst reports, return a
structured \texttt{Buy}, \texttt{Hold}, or \texttt{Sell} proposal with
reasoning and optional entry, stop, and natural-language sizing guidance.

\textbf{Initial directional and risk passes.} A bull researcher constructs one
evidence-based case from the market and news reports before the research
manager forms the investment plan. After the trader proposal, an aggressive
risk analyst performs one initial risk--reward pass before the portfolio
manager produces the final rating. With both configured debate-round counts set
to zero, these two initial passes still execute, while the bear, conservative,
and neutral follow-up turns are skipped.
\end{promptbox}

\begin{promptbox}{Final Decision Prompt for TradingAgents}
\small
As the Portfolio Manager, synthesize the research plan, trader proposal,
lessons from prior decisions, and any available risk-analysis history for
\texttt{\{instrument\}}. Select exactly one rating:
\texttt{Buy}, \texttt{Overweight}, \texttt{Hold},
\texttt{Underweight}, or \texttt{Sell}. Be decisive and ground the decision in
specific evidence from the supplied reports. Return a structured object with
\texttt{rating}, \texttt{executive\_summary},
\texttt{investment\_thesis}, and optional \texttt{price\_target} and
\texttt{time\_horizon}. The rating, rather than a model-generated quantity, is
passed to the deterministic position-mapping rule described above.
\end{promptbox}

In the deployed configuration, TradingAgents runs the market and news analysts,
one initial bull-research pass, the research manager and trader, one initial
aggressive-risk pass, and the portfolio manager. It disables the subsequent
bull--bear and three-way risk-debate rounds. The final decision prompt therefore
receives the research plan and trader proposal derived from the current reports,
the initial risk-analysis history, and prior lessons, but no follow-up debate
turns.


\newtcolorbox{finmemprompt}[1]{
  enhanced,
  breakable,
  colback=white,
  colframe=black!35,
  colbacktitle=black!32,
  coltitle=white,
  fonttitle=\bfseries\large,
  title={#1},
  boxrule=0.55pt,
  arc=2pt,
  left=12pt,
  right=12pt,
  top=10pt,
  bottom=10pt,
  toptitle=5pt,
  bottomtitle=5pt,
  before skip=9pt,
  after skip=9pt
}

\newcommand{\finmemsentiment}{%
In investment decision-making, analyzing financial sentiment is pivotal,
offering insights into market perceptions and forecasting potential market
trends. Sentiment analysis divides market opinions into three categories:
positive, negative, and neutral. A positive sentiment signals optimism about
future prospects, often leading to increased buying activity, while negative
sentiment reflects pessimism, likely causing selling pressures. Neutral
sentiment indicates either uncertainty or a balanced view, suggesting that
investors are neither overly bullish nor bearish. Leveraging these sentiment
indicators enables investors and analysts to fine-tune their strategies to
better match the prevailing market atmosphere. Additionally, news about
competitors can significantly impact a company's cryptocurrency price. For
example, if a competitor unveils a groundbreaking product, it may lead to a
decline in the cryptocurrency prices of other companies within the same
industry as investors anticipate potential market share losses.%
}

\newcommand{\finmemmomentum}{%
The information below provides a summary of cryptocurrency price fluctuations
over the previous few days, which is the ``Momentum'' of a cryptocurrency. It
reflects the trend of a cryptocurrency. Momentum is based on the idea that
securities that have performed well in the past will continue to perform well,
and conversely, securities that have performed poorly will continue to perform
poorly.%
}

FinMem combines a structured response format with stage-specific instructions.
The system prompt below specifies a JSON response containing an investment
decision, a textual justification, and supporting memory identifiers grouped
by memory layer. These identifiers link the explanation to the retrieved
evidence, while the structured output supports downstream decision recording
and memory updates.

\begin{finmemprompt}{System Prompt for FinMem}\small
You are a financial trading assistant. You MUST respond STRICTLY and ONLY in
valid JSON format. The JSON object MUST include the following keys:
\texttt{investment\_decision}, \texttt{summary\_reason},
\texttt{short\_memory\_ids}, \texttt{mid\_memory\_ids},
\texttt{long\_memory\_ids}, \texttt{reflection\_memory\_ids}. The field
\texttt{investment\_decision} MUST be exactly one of: ``buy'', ``sell'',
``hold''. The field \texttt{summary\_reason} MUST be a non-empty string. The
memory id fields MUST be arrays of integers. If there is no evidence for a
memory field, output an empty array. Do not output markdown. Do not output any
text outside the JSON object.
\end{finmemprompt}

During warm-up, FinMem is given an observed next-day price-change signal from
historical data and asked to explain the outcome using the available memories.
This is a retrospective explanation task rather than a prediction of an
unobserved outcome. The generated explanation is stored in the reflection
memory for subsequent retrieval, allowing later decisions to draw on summaries
of earlier market observations.

\begin{finmemprompt}{Instruction for FinMem}\small
\emph{Braced fields are runtime substitutions; italic conditions are not sent
to the model. Empty memory layers and their headings are omitted.
Here \texttt{\{future\_record\}} is $(p_t-p_{t+1})/p_t$ in the implementation.}

The current date is \texttt{\{cur\_date\}}. Here are the observed financial
market facts: for \texttt{\{symbol\}}, the price difference between the next
trading day and the current trading day is: \texttt{\{future\_record\}}.

\texttt{\{short\_memory\_section\}}\par
\emph{If short-term memories are present:}\par
\finmemsentiment\par
\texttt{\{mid\_memory\_section\}}\par
\texttt{\{long\_memory\_section\}}\par
\texttt{\{reflection\_memory\_section\}}\par
\emph{If three-day momentum is nonzero:}\par
\finmemmomentum\par
The cumulative return of past 3 days is
\texttt{\{positive\_or\_negative\}}.

Given the following information, can you explain to me why the financial
market fluctuation from current day to the next day behaves like this?
Summarize the reason of the decision. Your should provide a summary
information and the id of the information to support your summary.
\end{finmemprompt}

During testing and paper trading, the instruction omits the next-day outcome
and requests a decision based on retrieved memories and historical momentum.
Empty memory layers are excluded; the sentiment passage is included when
short-term memories are present, and the three-day momentum passage is added
only when momentum is nonzero. The instruction explicitly encourages buying
when sentiment or momentum is positive and permits holding when the
buy--sell choice is unclear, while requiring memory identifiers to support
the decision.

\begin{finmemprompt}{Test and Paper-Trading Instruction}\small
\emph{Braced fields are runtime substitutions; italic conditions are not sent
to the model. Empty memory layers and their headings are omitted. The
constructor does not insert a numerical value for current holdings.}

The ticker of the cryptocurrency to be analyzed is \texttt{\{symbol\}} and
the current date is \texttt{\{cur\_date\}}.

\texttt{\{short\_memory\_section\}}\par
\emph{If short-term memories are present:}\par
\finmemsentiment\par
\texttt{\{mid\_memory\_section\}}\par
\texttt{\{long\_memory\_section\}}\par
\texttt{\{reflection\_memory\_section\}}\par
\emph{If three-day momentum is nonzero:}\par
\finmemmomentum\par
The cumulative return of past 3 days is
\texttt{\{positive\_or\_negative\}}.

Given the information, can you make an investment decision? Just summarize
the reason of the decision. please consider the mid-term information, the
long-term information, the reflection-term information only when they are
available. If there no such information, directly ignore the impact for
absence such information. please consider the available short-term information
and sentiment associated with them. please consider the momentum of the
historical cryptocurrency price. When momentum or cumulative return is
positive, you are a risk-seeking investor. When momentum or cumulative return
is negative, you are a risk-averse investor. In particular, you should choose
to 'buy' when the overall sentiment is positive or momentum/cumulative return
is positive. please consider how much shares of the cryptocurrency the
investor holds now. You should provide exactly one of the following investment
decisions: buy or sell. When it is very hard to make a 'buy'-or-'sell'
decision, then you could go with 'hold' option. You also need to provide the
ids of the information to support your decision.
\end{finmemprompt}




\end{document}